\documentclass[11pt,a4paper]{article}
\usepackage[table]{xcolor}
\usepackage[hyperref]{acl2021}
\usepackage{times}
\usepackage{latexsym}
\usepackage{xcolor}

\newcommand{\mycomment}[3]{}

\newcommand{\ignore}[1]{}
\usepackage{graphics}
\usepackage{graphicx}
\usepackage{rotating}
\usepackage{multirow}
\usepackage{adjustbox}

\usepackage{array,booktabs}
\usepackage{subcaption}
\usepackage{float}
\usepackage{amsmath}
\usepackage{amssymb}
\usepackage{pifont}%
\usepackage{bm}

\usepackage{microtype}

\aclfinalcopy 

\usepackage{array}
\usepackage{colortbl}

\usepackage{url}

\usepackage{pgfplots}
\usepackage{contour}
\usepackage{xspace}
\usepackage{paralist}
\pgfplotsset{width=10cm,compat=1.9} 

\usepackage{algorithm,algpseudocode}
\usepackage{amsmath}
\usepackage{amsfonts}
\usepackage{amssymb}

\definecolor{darkblue}{rgb}{0,0,.5}
\definecolor{darkgreen}{rgb}{0,.5,0}
\definecolor{lightgray}{rgb}{.8,.8,.8}
\definecolor{aliceblue}{rgb}{0.75, 0.75, 1.0}
\definecolor{darkseagreen}{rgb}{0.46, 0.74, 0.46}
\definecolor{alizarin}{rgb}{0.82, 0.1, 0.26}
\definecolor{airforceblue}{rgb}{0.36, 0.54, 0.66}
\definecolor{red_graph}{rgb}{0.98, 0.8, 0.8}
\definecolor{blue_graph}{rgb}{0.8, 0.98, 0.8}
\definecolor{red}{rgb}{0.8, 0.0, 0.0}

\newcommand{\divergentmbert}{Divergent m\textsc{bert}\xspace}
\newcommand{\eq}{\textsc{eq}\xspace}
\newcommand{\dive}{\textsc{div}\xspace}
\newcommand{\bleu}{\textsc{bleu}\xspace}
\newcommand{\meteor}{\textsc{meteor}\xspace}
\newcommand{\bpe}{\textsc{bpe}\xspace}
\newcommand{\laser}{\textsc{laser}\xspace}
\newcommand{\equivalents}{\textsc{equivalents}\xspace}
\newcommand{\divagnostic}{\textsc{div-agnostic}\xspace}
\newcommand{\divtagged}{\textsc{div-tagged}\xspace}
\newcommand{\divfactors}{\textsc{div-factorized}\xspace}
\newcommand{\ted}{\textsc{ted}\xspace}
\newcommand{\nmt}{\textsc{nmt}\xspace}
\newcommand{\fren}{\textsc{fr}$\rightarrow$\textsc{en}\xspace}
\newcommand{\enfr}{\textsc{en}$\rightarrow$\textsc{fr}\xspace}

\newcommand{\lexical}{\textsc{lexical substitution}\xspace}
\newcommand{\replacement}{\textsc{phrase replacement}\xspace}
\newcommand{\subtree}{\textsc{subtree deletion}\xspace}

\newcommand{\debugcolor}{white}

\newcommand{\rowlabel}[2]{\begin{tikzpicture}[scale=0.15]
\fill[\debugcolor] (0,0) rectangle (0.1,4);
\fill[\debugcolor] (0,0) rectangle (0.1,-#1);

\draw (0,0) node[anchor=south] {\bf #2}; 
\end{tikzpicture}}
\newcommand{\rowlabelx}[4]{\begin{tikzpicture}[scale=0.15]
\fill[\debugcolor] (0,0) rectangle (0.1,4);
\fill[\debugcolor] (0,0) rectangle (0.1,-#1);
\draw (0,+1.5) node[anchor=south] {\bf \hspace{-#4in}  #2}; 
\draw (0.0,-1.5) node[anchor=south] {\bf #3}; 
\end{tikzpicture}}
\newcommand{\barchart}[3]{\begin{tikzpicture}[scale=0.17]
\fill[\debugcolor] (5.7,0) rectangle (5.8,-#1);
\fill[\debugcolor] (5.7,0) rectangle (5.8,4);
\fill[pink] (0,0) rectangle (5.5,0#3);
\draw (2.75,0) node[anchor=south] {#2};
\draw (5.75,0) node[anchor=south] {};
\draw (2.75,0) node[anchor=north] {\textcolor{red}{#3}};
\end{tikzpicture}}
\newcommand{\parchart}[3]{\begin{tikzpicture}[scale=0.17]
\fill[\debugcolor] (5.7,0) rectangle (5.8,-#1);
\fill[\debugcolor] (5.7,0) rectangle (5.8,4);
\fill[aliceblue] (0,0) rectangle (5.5,0#3);
\draw (2.75,0) node[anchor=south] {#2};
\draw (5.75,0) node[anchor=south] {};
\draw (2.75,0) node[anchor=north] {\textcolor{blue}{#3}};
\end{tikzpicture}}
\newcommand{\graydt}[1]{\begin{tikzpicture}[scale=0.7]
\fill[lightgray] (0,0) rectangle (#1,0.4);
\end{tikzpicture}}
\newcommand{\combdt}[2]{\begin{tikzpicture}[scale=0.7]
\fill[airforceblue] (0,0) rectangle (#1,0.4);
\fill[alizarin] (#1,0) rectangle (#1+#2,0.4);
\end{tikzpicture}}

\contourlength{0.2pt}
\newcommand{\hz}{\vphantom{\parbox[c]{0.08cm}{\rule{0.08cm}{0.19cm}}}}
\newcommand{\ua}{%
  \colorbox{lightgray}{\hz\tiny{$\uparrow$}}%
}
\newcommand{\ph}{%
  \colorbox{white}{\hz\color{white}\tiny{$\downarrow$}}%
}
\newcommand{\verticalchart}[2]{\begin{tikzpicture}[scale=0.17]
\fill[color=lightgray] (0.1,0.1) rectangle (5.5,3*#2);
\draw[color=gray] (0.1,0.1) rectangle (5.5,3*#2);
\draw (5.75,0) node[anchor=south] {};
\draw (2.75,0) node[anchor=north] {\textcolor{gray}{#2}};
\end{tikzpicture}}
\newcommand{\verticalpairchart}[3]{\begin{tikzpicture}[scale=0.17]
\fill[gray] (0.1,0.1) rectangle (5.5,3*#2);
\draw[black] (0.1,0.1) rectangle (5.5,3*#2);
\fill[gray] (6.1,0.1) rectangle (11.6,3*#3);
\draw[pattern=grid, pattern color=black] (6.1,0.1)  rectangle (11.6,3*#3);
\draw (5.75,0) node[anchor=south] {};
\draw (2.75,0) node[anchor=north] {\textcolor{gray}{#2}};
\draw (8.75,0) node[anchor=north] {\textcolor{gray}{#3}};
\end{tikzpicture}}

\title{Beyond Noise: Understanding the Impact of Fine-grained \\Semantic Divergences on Neural Machine Translation}
\title{Beyond Noise: Mitigating the Impact of Fine-grained \\Semantic Divergences on Neural Machine Translation}

\author{Eleftheria Briakou \normalfont{and}  \textbf{Marine Carpuat} \\
  Department of Computer Science \\
  University of Maryland\\
  College Park, MD $20742$, USA\\
  \texttt{\href{mailto:ebriakou@cs.umd.edu}{ebriakou@cs.umd.edu}, \href{mailto:marine@cs.umd.edu}{marine@cs.umd.edu}}} 
\date{}
\begin{document}
\maketitle
\usetikzlibrary{patterns}

\begin{abstract}
While it has been shown that Neural Machine Translation (\textsc{nmt}) is highly sensitive to noisy parallel training samples, prior work treats all types of mismatches between source and target as noise. As a result, it remains unclear how samples that are mostly equivalent but contain a small number of semantically divergent tokens impact \nmt training. To close this gap, we analyze the impact of different types of fine-grained semantic divergences on Transformer models. We show that models trained on synthetic divergences output degenerated text more frequently and are less confident in their predictions. Based on these findings, we introduce a divergent-aware \textsc{nmt} framework that uses factors to help \textsc{nmt} recover from the degradation caused by naturally occurring divergences, improving both translation quality and model calibration on \textsc{en}$\leftrightarrow$\textsc{fr} tasks.
\end{abstract}

\section{Introduction}\label{sec:introduction}
\begin{figure}[t!]
    \centering
    \includegraphics[scale=0.355]{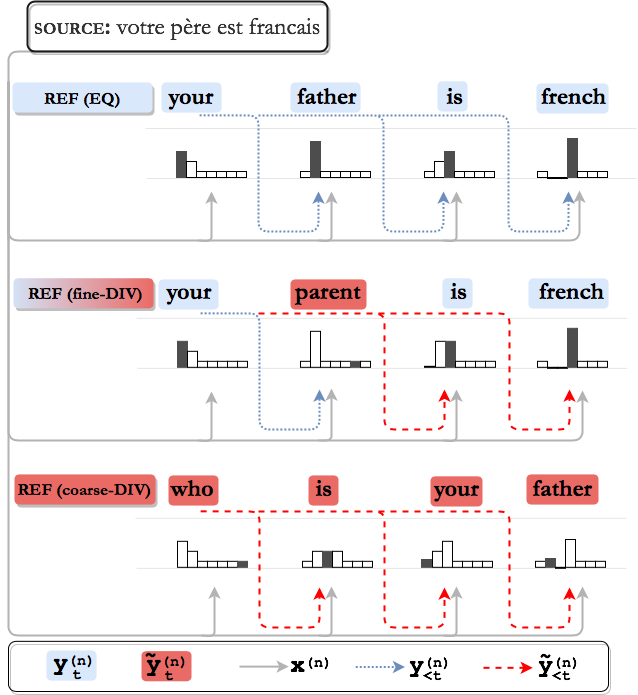}
    \caption{Equivalent vs. Divergent references on \nmt training. Fine-grained divergences (i.e., \textsc{ref} (fine-\textsc{div})) provide an imperfect yet potentially useful signal depending on the time step $t$. 
    }\vspace{-1em}
    \label{fig:nmt_interactions}
\end{figure}

While parallel texts are essential to Neural Machine Translation (\nmt), the degree of parallelism varies widely across samples in practice, for reasons ranging from noise in the extraction process~\cite{roziewski-stokowiec-2016-languagecrawl} to non-literal translations~\cite{zhai19,zhai-etal-2020-detecting}. 
For instance (Figure~\ref{fig:nmt_interactions}), a French \textsc{source} could be paired with an exact translation into English~(\eq), with a mostly equivalent translation where only a few tokens convey divergent meaning~(fine-\dive), or with a semantically unrelated, noisy reference~(coarse-\dive).
Yet, prior work treats parallel samples in a binary fashion: coarse-grained divergences are viewed as noise to be excluded from training~\cite{koehn-etal-2018-findings}, whilst others are typically regarded as gold-standard equivalent translations. As a result, the impact of fine-grained divergences on \nmt remains unclear.

This paper aims to understand and mitigate the impact of fine-grained semantic divergences in \nmt. We first contribute an \textbf{analysis} of how fine-grained divergences in training data affect \nmt quality and confidence.  Starting from a set of equivalent English-French WikiMatrix sentence pairs, we simulate divergences by gradually ``corrupting'' them with synthetic \textit{fine-grained divergences}. Following \citet{khayrallah-koehn-2018-impact}---who, in contrast, study the impact of \textit{noise} on \textsc{mt}---we control for different \textit{types} of fine-grained semantic divergences and different \textit{ratios} of equivalent vs. divergent data. Our findings indicate that these imperfect training references: hurt translation quality (as measured by \bleu and \meteor) once they overwhelm equivalents; output degenerated text more frequently; 
and increase the uncertainty of models' predictions.

Based on these findings, we introduce a \textbf{divergent-aware \nmt framework} that incorporates information about which tokens are indicative of semantic divergences between the source and target side of a training sample.
Source-side divergence tags are integrated as feature factors~\cite{fsmti,sennrich-haddow-2016-linguistic,hoang-etal-2016-improving}, while target-side divergence tags form an additional output sequence generated in a multi-task fashion~\cite{factors_nmt,garcia2018}. 
Results on \textsc{en}$\leftrightarrow$\textsc{fr} translation show that our approach is a successful \textbf{mitigation strategy}: it helps \nmt recover from the negative impact of fine-grained divergences on translation quality, with fewer degenerated hypotheses, and more confident and better calibrated predictions.  We make our code publicly available: \url{https://github.com/Elbria/xling-SemDiv-NMT}.

%
%
\section{Background \& Motivation}\label{sec:background_and_motivation}
\paragraph{Cross-lingual Semantic Divergences} We use this term to refer to meaning differences in aligned bilingual text~\cite{vyas-etal-2018-identifying,carpuat-etal-2017-detecting}.
Divergences in manual translation might arise due to the translation process~\cite{zhai-etal-2018-construction} and result in non-literal translations~\cite{zhai-etal-2020-detecting}. Divergences might also arise in parallel text extracted from multilingual comparable resources. For instance, in Wikipedia, documents aligned across languages might contain parallel segments that share important content, yet they are not perfect translations of each other, yielding fine-grained semantic divergences~\cite{smith-etal-2010-extracting}. Finally coarse-grained divergences might result from the process of automatically mining and aligning corpora from monolingual data \cite{fung-cheung-2004-multi,Munteanu:2005:IMT:1110825.1110828}, or web-scale parallel text \cite{smith-etal-2013-dirt,searching,espla-etal-2019-paracrawl}.
\paragraph{Noise vs. Semantic Divergences} In the context of \textsc{mt}, noise often refers to mismatches in \textbf{web-crawled} parallel corpora that are collected without guarantees about their quality. \citet{khayrallah-koehn-2018-impact} define five frequent types of noise found
in the German-English Paracrawl corpus: \textit{misaligned sentences}, \textit{disfluent text}, \textit{wrong language}, \textit{short segments}, and \textit{untranslated sentences}. They examine the impact of noise on translation quality and find that untranslated training instances cause \nmt models to copy the input sentence at inference time. Their findings motivated a shared task dedicated to filtering noisy samples from web-crawled data at \textsc{wmt}, since 2018~\cite{koehn-etal-2018-findings, koehn-etal-2019-findings, koehn-etal-2020-findings}. 
This work moves beyond such coarse divergences and focuses instead on fine-grained divergences that affect a small number of tokens within mostly equivalent pairs and that can be found even in high-quality parallel corpora.

\paragraph{Training Assumptions}
\nmt models are typically trained to maximize the log-likelihood of the training data, $\mathcal{D} \equiv \{ (\bm{x}^{(n)}, \bm{y}^{(n)}) \}_{n=1}^{N}$, where ($\bm{x}^{(n)}, \bm{y}^{(n)}$) is the $n$-th sentence pair consisting of sentences that are \textbf{assumed to be translations of each other}. Under this assumption, model parameters are updated to maximize the \textbf{token-level} cross-entropy loss: 
\begin{equation}
    \mathcal{J}(\theta) = \sum_{n=1}^{N} \sum_{t=1}^{T} \log p(y_t^{(n)} \mid \bm{y}_{<t}^{(n)}, \bm{x}^{(n)}; \theta)
\end{equation}

In Figure~\ref{fig:nmt_interactions}, we illustrate how semantic divergences interact with \nmt training.  In the case of coarse divergences, both the prefixes $\widetilde{\bm{y}}_{t<1}^{(n)}$ and targets $\widetilde{y}_{t}^{(n)}$,
yield a noisy training signal at each time step $t$, which motivates excluding them from the training pool entirely. In the case of fine-grained divergences, the assumption of \textit{semantic equivalence} is only partially broken. Depending on the time step $t$, we might thus condition the prediction of the next token on partially corrupted prefixes, encourage the model to make a wrong prediction, or do a combination of the above. This suggests that fine-grained divergent samples provide a noisy yet potentially useful training signal depending on the time step. Meanwhile, fine-grained divergences increase uncertainty in the training data, and as a result might impact models' confidence in their predictions, as noisy untranslated samples do \citep{ott2018analyzing}. This work seeks to clarify and mitigate their impact on \nmt, accounting for both translation quality and model confidence.
%

%

%
\section{Analyzing the Impact of Divergences}\label{sec:analysis_on_synthetic}
\subsection{Method}\label{sec:a1}
We evaluate the impact of semantic divergences on \textsc{nmt} by injecting increasing amounts of synthetic divergent samples during training, following the methodology of \citet{khayrallah-koehn-2018-impact} for noise. We focus on three types of divergences, which were found to be frequent in parallel corpora. They are fine-grained as they represent discrepancies between the source and target segments at a word or phrase level: \textbf{\lexical} aims at mimicking \textit{particularization} and \textit{generalization} operations resulting from non-literal translations~\cite{zhai-etal-2019-hybrid,zhai-etal-2020-building}; \textbf{\replacement} mimics phrasal mistranslations; \textbf{\subtree} simulates missing phrasal content from the source or target side.

Synthetic divergent samples are automatically generated by corrupting semantically equivalent sentence pairs, following the methodology introduced by~\citet{briakou-carpuat-2020-detecting}. Equivalents are identified by their \divergentmbert classifier that yields an F$1$ score of $84$, on manually annotated WikiMatrix data, despite being trained on synthetic data.
For \lexical we corrupt equivalents by substituting words with their hypernyms or hyponyms from WordNet, for \replacement we replace sequences of words with phrases of matching \textsc{pos} tags, and for \subtree we randomly delete subtrees in the dependency parse tree of either the source or the target. Having access to those $4$ versions of the same corpus (one initial equivalent and three synthetic divergences), we mix equivalents and divergent pairs introducing one type of divergence at a time (corpora statistics are included in~\ref{sec:synthetic_divergences_stats}). Finally, we evaluate the translation quality and uncertainty of the resulting translation models.

\subsection{Experimental Set-Up}\label{sec:a2}
\paragraph{Training Data} We train our models on the parallel WikiMatrix French-English corpus~\cite{wikimatrix}, which consists of sentence pairs mined from Wikipedia pages using language-agnostic sentence embeddings (\laser)~\cite{Artetxe2019MassivelyMS}. Previous annotations show that $40\%$ of sentence pairs in a random sample contain fine-grained divergences \cite{briakou-carpuat-2020-detecting}.

After cleaning noisy samples using simple rules (i.e., exclude pairs that are a) too short or too long, b) mostly numbers, c) almost copies based on edit distance), we extract \textit{equivalent} samples using the \divergentmbert model. Table~\ref{tab:wikimatrix_divs} presents statistics on the extracted pairs, along with the corpus created if we threshold the \laser score at $1.04$, as suggested by \citet{wikimatrix}.
\begin{table}[!ht]
    \centering
    \scalebox{0.99}{
    \begin{tabular}{lr}
    \rowcolor{gray!10}
    \textbf{Corpus} & \textbf{\#Sentences} \\
    \addlinespace[0.3em]
     \textsc{Wikimatrix}       & $6{,}562{,}360$   \\
     \addlinespace[0.2em]
     {\hskip 0.05in} \textsc{+ heuristic filtering} & $2{,}437{,}108$ \\
     \addlinespace[0.2em]
    {\hskip 0.2in} \textsc{+ laser filtering}       & $1{,}250{,}683$   \\
    \addlinespace[0.2em]
    {\hskip 0.2in} + divergentm\textsc{bert filtering}         & $751{,}792$    \\ 
    \addlinespace[0.2em]
    \end{tabular}}   
    \caption{WikiMatrix  \textsc{en-fr} corpus statistics.}
    \label{tab:wikimatrix_divs}
\end{table}
\paragraph{Development and Test data} We use the official development and test splits of the \ted corpus~\cite{qi-etal-2018-pre}, consisting of $4{,}320$ and $4{,}866$ gold-standard translation pairs, respectively. All models share the same \bpe vocabulary. We average results across runs with $3$ different random seeds.
\paragraph{Preprocessing} We use the standard Moses scripts \cite{koehn-etal-2007-moses} for punctuation normalization, true-casing, and tokenization. We learn $32$K \textsc{bpe}s~\cite{sennrich-etal-2016-neural} using SentencePiece~\cite{kudo-richardson-2018-sentencepiece}.  
\paragraph{Models} We use the base Transformer architecture \cite{vaswani}, with embedding size of $512$, transformer hidden size of $2{,}048$, $8$ attention heads, $6$ transformer layers, and dropout of $0.1$. 
Target embeddings are tied with the output layer weights.
We train with label smoothing ($0.1$). We optimize with Adam~\cite{Kingma2015AdamAM} with a batch size of $4{,}096$ tokens and checkpoint models every $1{,}000$ updates. The initial learning rate is $0.0002$, and it is reduced by $30$\% after $4$ checkpoints without validation perplexity improvement. We stop training after $20$ checkpoints without improvement. We select the best checkpoint based on validation \bleu~\cite{papineni-etal-2002-bleu}. All models are trained on a single GeForce \textsc{gtx} $1080$ \textsc{gpu}.

\subsection{Findings}\label{sec:a3}
\paragraph{Translation Quality}
Table~\ref{tab:bleu_impact} presents the impact of semantic divergences on \bleu and \meteor. Corrupting equivalent bitext with fine-grained divergences hurts translation quality across the board. In most cases, the degradation is proportional to the percentage of corrupted training samples. \lexical causes the largest degradation for both metrics. The degradation is relatively smaller for \meteor than \bleu, which we attribute to the fact that \meteor allows matches between synonyms when comparing references to hypotheses. \subtree and \lexical corruptions lead to significant degradation at~$\geq 50\%$ (\bleu; standard deviations across reruns are $<0.4$). By contrast,  Transformers are more robust to \replacement corruptions, as degradations are only significant after corrupting~$\geq 70\%$ (\bleu) of equivalents.
\begin{table*}[!ht]
    \scalebox{0.7}{
    \begin{tabular}{l@{\hskip 0.2in}cccccccccccc}

    & & \multicolumn{5}{c}{\bleu} &  \multicolumn{5}{c}{\meteor} \\
    \cmidrule(lr){2-7}  \cmidrule(lr){8-13}
    \addlinespace[0.3em]

    & $\mathbf{0\%}$ &  $\mathbf{10\%}$ & $\mathbf{20\%}$ & $\mathbf{50\%}$ & $\mathbf{70\%}$ & $\mathbf{100\%}$ &   $\mathbf{0\%}$ &  $\mathbf{10\%}$ 
    & $\mathbf{20\%}$ & $\mathbf{50\%}$ & $\mathbf{70\%}$ & $\mathbf{100\%}$ \\ 
 
    \addlinespace[0.3em]
    \cmidrule(lr){2-7}
    \cmidrule(lr){8-13}
    \addlinespace[0.3em]

    \rowlabelx{4.5}{\textsc{phrase}}{\textsc{replacement}}{35} & 
        \barchart{5.5}{$30.89$}{+0.00} & 
        \barchart{5.5}{$31.00$}{+0.11} & 
        \barchart{5.5}{$30.82$}{-0.07} & 
        \barchart{5.5}{$30.40$}{-0.49} & 
        \barchart{5.5}{$29.74$}{-1.15} & 
        \barchart{5.5}{$27.01$}{-3.88} &
        \parchart{5.5}{$33.74$}{+0.00} & 
        \parchart{5.5}{$33.63$}{-0.11} &
        \parchart{5.5}{$33.66$}{-0.08} & 
        \parchart{5.5}{$33.54$}{-0.20} & 
        \parchart{5.5}{$33.12$}{-0.62} & 
        \parchart{5.5}{$31.02$}{-2.72}\\
    
      \cmidrule(lr){2-7}
      \cmidrule(lr){8-13}
      \addlinespace[0.3em]

    \rowlabelx{4.5}{\textsc{subtree}}{\textsc{deletion}}{8} & 
        \barchart{5.5}{$30.89$}{+0.00} & 
        \barchart{5.5}{$30.80$}{-0.09} & 
        \barchart{5.5}{$30.62$}{-0.27} & 
        \barchart{5.5}{$28.95$}{-1.94} & 
        \barchart{5.5}{$29.00$}{-1.89} & 
        \barchart{5.5}{$27.50$}{-3.39} &
        \parchart{5.5}{$33.74$}{+0.00} & 
        \parchart{5.5}{$33.61$}{-0.13} &
        \parchart{5.5}{$33.38$}{-0.36} & 
        \parchart{5.5}{$32.17$}{-1.57} & 
        \parchart{5.5}{$32.09$}{-1.65} &    
        \parchart{5.5}{$31.44$}{-2.30}\\

        \cmidrule(lr){2-7}
        \cmidrule(lr){8-13}
        \addlinespace[0.3em]

    \rowlabelx{4.5}{\textsc{lexical}}{\textsc{substitution}}{30} & 
        \barchart{5.5}{$30.89$}{+0.00} &
        \barchart{5.5}{$30.72$}{-0.17} & 
        \barchart{5.5}{$30.49$}{-0.40} & 
        \barchart{5.5}{$25.04$}{-5.85} & 
        \barchart{5.5}{$26.57$}{-4.32} & 
        \barchart{5.5}{$25.18$}{-5.71} &
        \parchart{5.5}{$33.74$}{+0.00} & 
        \parchart{5.5}{$33.56$}{-0.18} &
        \parchart{5.5}{$33.50$}{-0.24} & 
        \parchart{5.5}{$29.59$}{-4.15} & 
        \parchart{5.5}{$31.58$}{-2.16} & 
        \parchart{5.5}{$30.75$}{-2.99}\\
        
    \addlinespace[0.3em]
    \cmidrule(lr){2-7}
    \cmidrule(lr){8-13}
    \addlinespace[0.3em]

\end{tabular}}
\caption{Results for \fren translation on the \ted test set (means of $3$ runs). 
Bars denote degradation over \equivalents (i.e., $0\%$) across different $\%$ of corruption. Divergences hurt \bleu and \meteor when they overwhelm the training data. Transformers are particularly sensitive to fine nuances introduced by \lexical.
}
\label{tab:bleu_impact}
\end{table*}
\begin{figure*}[!t]
\begin{subfigure}{.5\textwidth}
  \centering
  \includegraphics[scale=0.45]{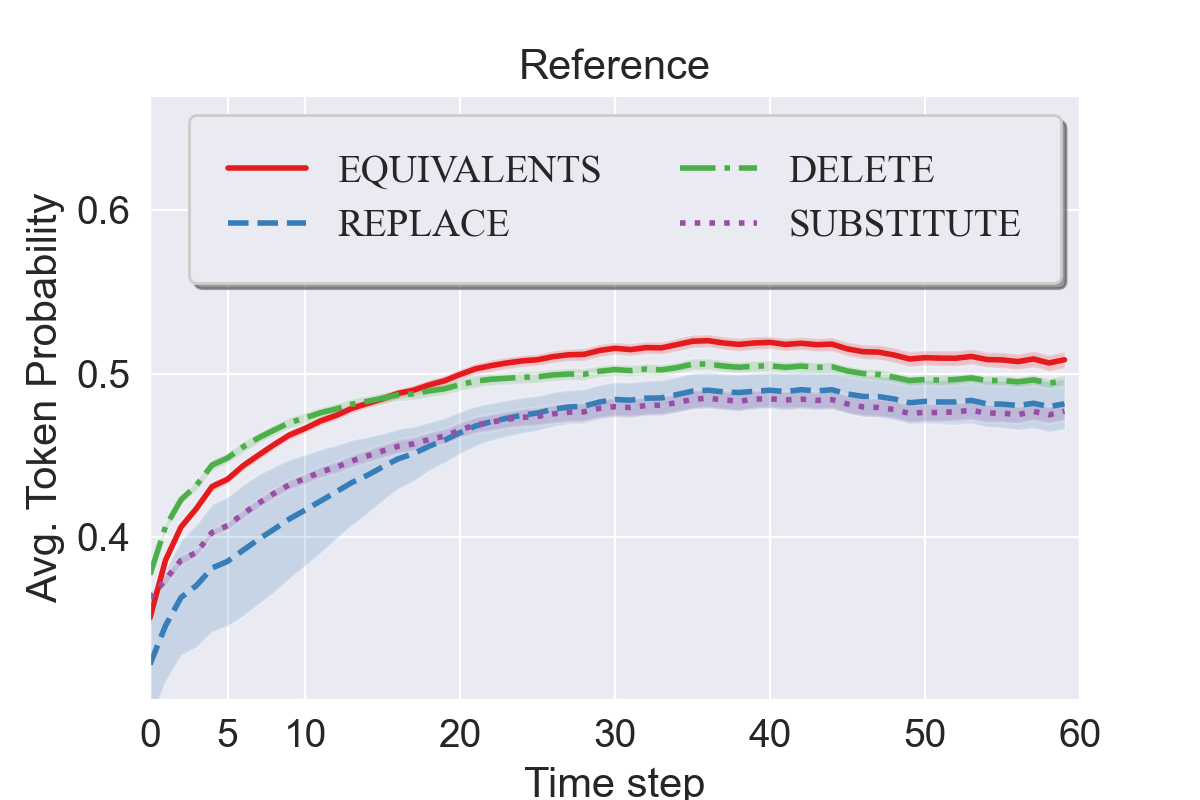}
\end{subfigure}
\begin{subfigure}{.5\textwidth}
  \centering
  \includegraphics[scale=0.45]{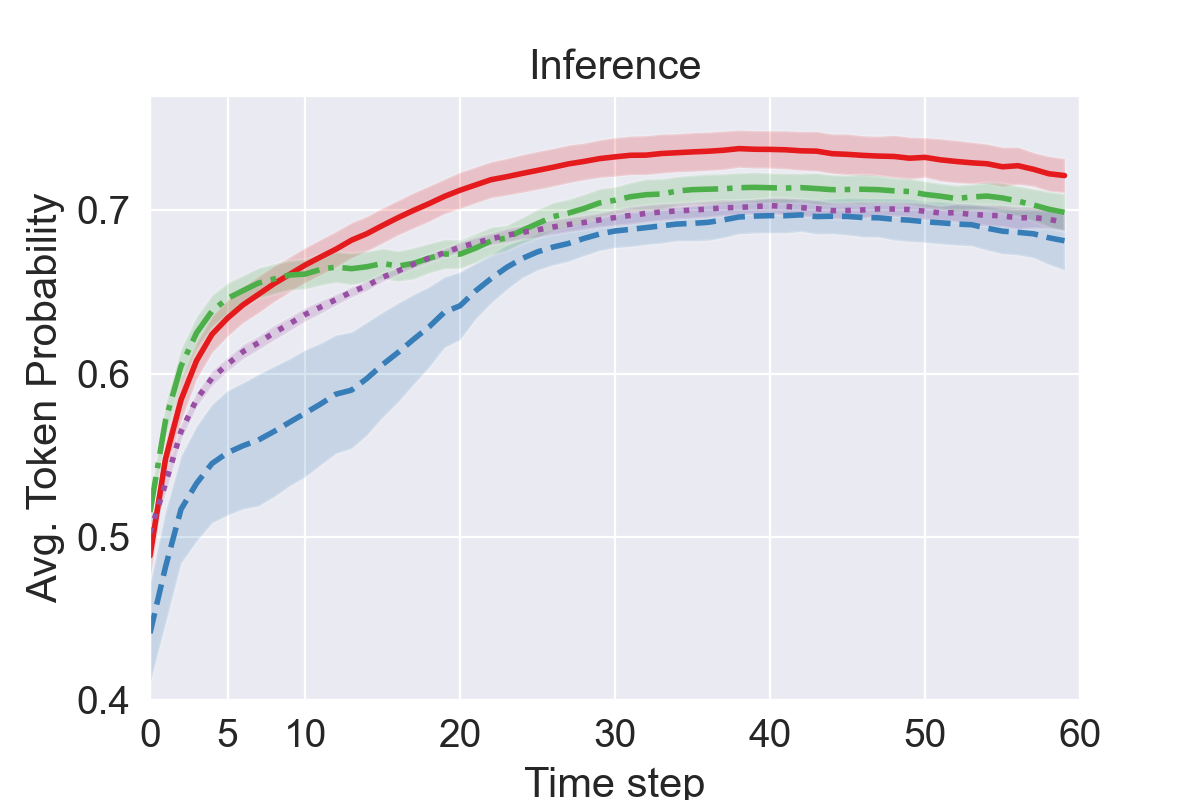}
\end{subfigure}
\caption{Average token probabilities of predictions conditioned on gold references (left) and beam search ($5$) prefixes (right).
Training on fine-grained divergences ($100\%$ corruption) increase \nmt model's uncertainty.}
\label{fig:uncertainty}
\end{figure*}
\begin{figure}[!ht]
    \centering
    \includegraphics[scale=0.45]{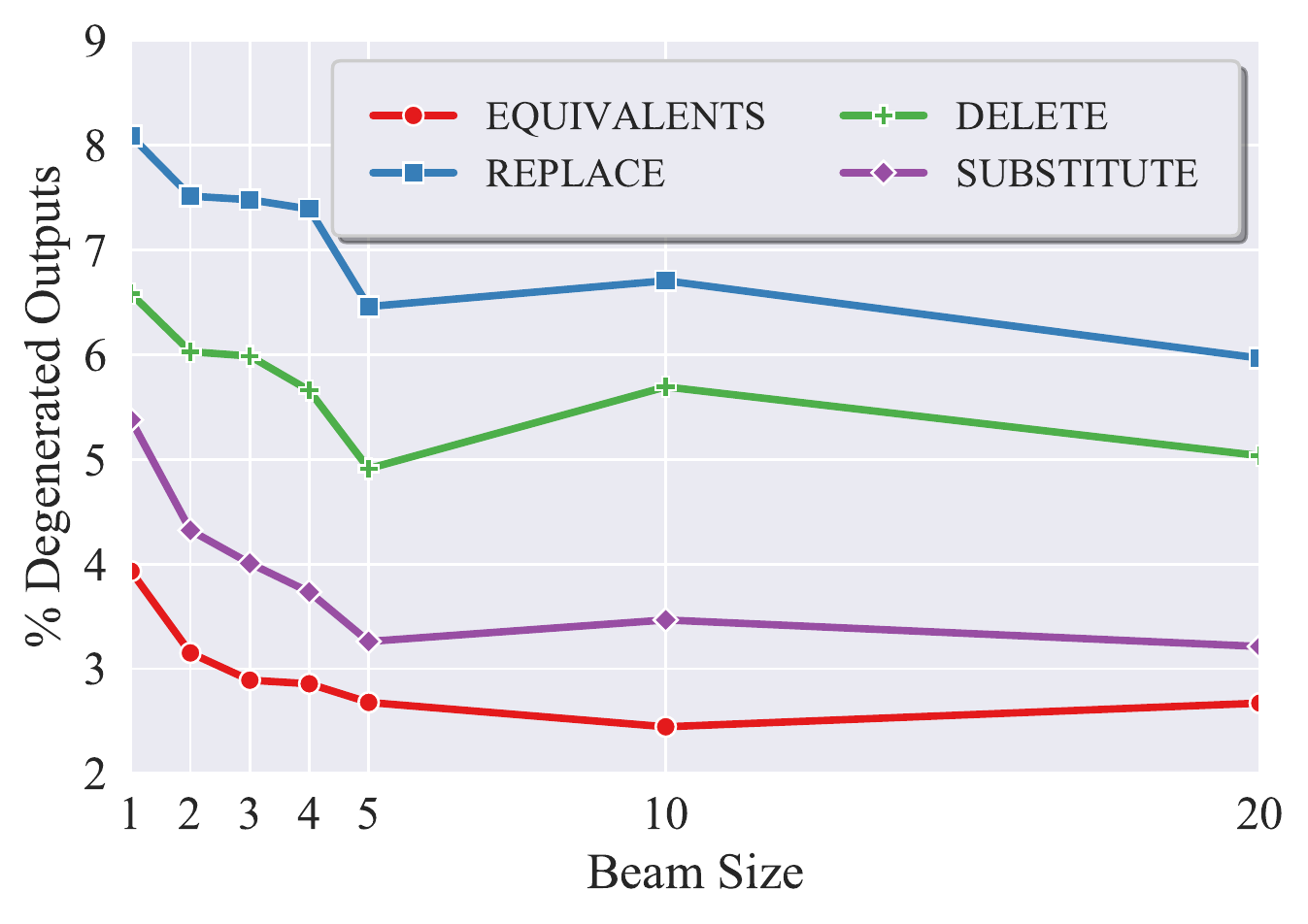}
    \caption{$\%$ of degenerated outputs as a function of beam size. \nmt training on fine-grained divergences ($100\%$ corruptions) increase the frequency of degenerated hypotheses across beams.}
    \label{fig:degenerated}\vspace{-1em}
\end{figure}
\paragraph{Token Uncertainty} We measure the impact of divergences on model uncertainty at training time and at test time. For the first, we extract the probability of a reference token conditioned on reference prefixes at each time step. For the latter, we compute the probability of the token predicted by the model given its own history of predictions. Figure~\ref{fig:uncertainty} shows that models trained on \equivalents  are more confident in their token level predictions both at inference and training time. 
\subtree mismatches affect models' confidence less than other types, while \replacement hurts confidence the most both at inference and at training time. Finally, we observe that differences across divergence types are larger in early decoding steps, while at later steps, they all converge below the \equivalents.
\paragraph{Degenerated Hypotheses}  When models are trained on $50\%$ or more divergent samples, the total length of their hypotheses is longer than the references. Manual analysis on models trained with $100\%$ of divergent samples suggests that this length effect is partially caused by  \textit{degenerated} text. Following \citet{degenerated}---who study this phenomenon for unconditional text generation---we define \textit{degenerations} as ``output text that is bland, incoherent, or gets stuck in repetitive loops''.\footnote{For instance, ``I've never studied sculpture, engineering and architecture, \textbf{and the engineering and architecture".}}

We automatically detect degenerated text in model outputs by checking whether they contain repetitive loops of $n$-grams that do not appear in the reference (details on the algorithm are in \ref{sec:AppendixA3}).
Figure~\ref{fig:degenerated} shows that exposing \nmt to divergences increases the percentage of degenerated outputs. Even with large beams, the models trained on divergent data yield more repetitions than the \equivalents. Moreover, divergences due to phrasal mismatches (\replacement and \subtree) yield more frequent repetitions than token-level mismatches (\lexical). Interestingly, the latter almost matches the frequency of repetitions in \equivalents with larger beams ($\geq5$).

\paragraph{Summary} Synthetic divergences hurt translation quality, as expected. More surprisingly, our study also reveals that this degradation is partially due to more frequent degenerated outputs, and that divergences impact models' confidence in their predictions. Different types of divergences have different effects: \lexical causes the largest degradation in translation quality, \subtree and \replacement increase the number of degenerated beam hypotheses, while \replacement also hurts the models' confidence the most.
Nevertheless, the impact of divergences on \bleu appears to be smaller than that of noise~\cite{khayrallah-koehn-2018-impact}.\footnote{While the absolute scores are not directly comparable across settings, \citet{khayrallah-koehn-2018-impact} report that noise has a more striking impact of $-8$ to $-25$ BLEU.} This suggests that noise filtering techniques are suboptimal to deal with fine-grained divergences.
%

%
\section{Mitigating the Impact of Fine-grained Divergences}\label{sec:mitigating_impact}
%
We now turn to naturally occurring divergences in WikiMatrix. We will see that their impact on model quality and uncertainty is consistent with that of synthetic divergences (\S~\ref{sec:b3}). We propose a divergent-aware framework for \textsc{nmt}  (\S~\ref{sec:b1}) that successfully mitigates their impact (\S~\ref{sec:b3}). 
%
\subsection{Factorizing Divergences for \nmt}\label{sec:b1}
We use \textbf{semantic factors} to inform \nmt of tokens that are indicative of meaning differences in each sentence pair. We tag divergent source and target tokens in parallel segments as equivalent (\eq) or divergent (\dive) using an m\textsc{bert}-based classifier
 trained on synthetic data. The classifier has a $45$ F$1$ score on a fine-grained divergence test set \citep{briakou-carpuat-2020-detecting}. The predicted tags are thus noisy, as expected on this challenging task, yet we will see that they are useful. An example is illustrated below:
\begin{tabular}{p{0.95\columnwidth}}
\\
\multirow{2}{*}{\textsc{src}}
\hspace{0.5em} \textsc{tokens} \hspace{0.6em} \it votre \color{red}{père} \color{black}{est francais} \\
\hspace{2.5em} \textsc{factors} \hspace{0.4em}  \eq \ \ \color{red}{\dive} \ \ \color{black}{\eq \ \  \eq}\\
\\
\multirow{2}{*}{\textsc{tgt}}
\hspace{0.5em} \textsc{tokens} \hspace{0.6em} \it your \color{red}{parent} \color{black}{is french} \\
\hspace{2.5em} \textsc{factors} \hspace{0.4em}  \eq \ \ \color{red}{\dive} \ \ \color{black}{\eq \ \  \eq}
\\
\end{tabular}
\paragraph{Source Factors} We follow \citet{sennrich-haddow-2016-linguistic} who 
represent the encoder input as a combination of token embeddings and linguistic features. Concretely, we look up separate embeddings vectors for tokens and source-side divergent predictions, which are then concatenated. The length of the concatenated vector matches the total embedding size.
\paragraph{Target Factors} Target-side divergence tags are an additional output sequence, as in \citet{factors_nmt}. At each time step the model produces two distributions: one over the token target vocabulary and one over the target factors. The model is trained to minimize a divergent-aware loss (Equation~\ref{eq:factorized_loss}). Terms in red (also, underlined) correspond to modifications to the traditional \nmt loss. At time step $t$, the model is rewarded to match the reference target 
$y_t^{(n)}$, conditioned on the source sequence of tokens ($\bm{x}^{(n)}$),  the source factors ($\bm{\omega}^{(n)}$), the token target prefix ($\bm{y}_{<t}^{(n)}$),  and the target factors prefix ($\bm{z}_{<t}^{(n)}$).  At the same time ($t$), the model is rewarded to match the factored predictions for the previous time step $\tau=t-1$. The time shift between the two target sequences is introduced
so that the model learns to firstly predict the reference token at $\tau$ and then its corresponding \eq vs. \dive label, at the same time step. The factored predictions are conditioned again on $\bm{x}^{(n)}$, $\bm{\omega}^{(n)}$, the target factor prefix  $\bm{z}_{<\tau}^{(n)}$ and the token prefix ($\bm{y}_{\leq \tau}^{(n)}$). 
\begin{equation}\label{eq:factorized_loss}
    \small
    \begin{split}
    \mathcal{L} =  -
    \sum_{n=1}^{N} \Bigg( & \underbrace{ \sum_{t=1}^{T}  \log p(y_t^{(n)} \mid \bm{y}_{<t}^{(n)}, \underline{{\color{red}{ \bm{z}_{<t}^{(n)}}}},  \bm{x}^{(n)}, \underline{{\color{red}{ \bm{\omega}^{(n)}}}}; \theta)}_{\tilde{\mathcal{L}}_{\textsc{mt}}^{(n)}} \\
    + &  
    \underbrace{\sum_{\tau=t-1}^{T} \log p( z_\tau^{(n)} \mid \bm{z}_{<\tau}^{(n)}, \bm{y}_{\leq \tau}^{(n)}, \bm{x}^{(n)}, \bm{\omega}^{(n)}; \theta)}_{ \underline{{\color{red}{ \mathcal{L}_{factor}^{(n) }}}} } \Bigg)
    \end{split}
\end{equation}
\paragraph{Inference} At test time, input tokens are tagged with \eq to encourage the model to predict an equivalent translation. We decode using beam search for predicting the translation sequence.
The token predictions are conditioned on both the token and the factors prefixes. The factor prefixes are greedily decoded and thus do not participate in beam search.

\subsection{Experimental Set-Up}\label{sec:b2}
\paragraph{Divergences} We conduct an extensive comparison of models exposed to different amounts of equivalent and divergent WikiMatrix samples. Starting from the pool of examples identified as divergent at \S\ref{sec:a2}, we rank and select the most fine-grained divergences by thresholding the \texttt{bicleaner} score~\cite{ramirez-sanchez-etal-2020-bifixer} at $0.5$, $0.7$ and $0.8$. For details, see \ref{sec:appendixA1}.
\paragraph{Models} We compare the factored models (\textbf{\divfactors}) for incorporating divergent tokens (\S\ref{sec:b1}) against:
\begin{inparaenum}
\item \textbf{\laser} models are trained on WikiMatrix pairs with a \laser score greater than $1.04$ \---\ the noise filtering strategy recommended by \citet{wikimatrix}. 
Our prior work shows that thresholding \laser might introduce a number of divergent data in the training pool varying from fine to coarse mismatches~\cite{briakou-carpuat-2020-detecting}.
\item \textbf{\equivalents} models are trained on WikiMatrix pairs detected as exact translations (\S\ref{sec:a2});
\item \textbf{\divagnostic} models are trained on equivalent and fine-grained divergent data without incorporating information that distinguishes between them;
\item \textbf{\divtagged} 
models distinguish equivalences from divergences by appending $<$\textsc{eq}$>$ vs.\ $<$\textsc{div}$>$ tags as source-side constraints~\cite{sennrich-etal-2016-controlling}.
\end{inparaenum}
\paragraph{Models' details} Our models are implemented in the \texttt{Sockeye2} toolkit~\cite{domhan-etal-2020-sockeye}.\footnote{\url{https://github.com/awslabs/sockeye}} We set the size of factor embeddings to $8$, the source token embeddings to $504$ and target embeddings to $514$, yielding equal model sizes across experiments. All other parameters are kept the same across models, as discussed in \S\ref{sec:a2}, except that target embeddings are not tied with output layer weights for factored models. More details are included in~\ref{sec:appendixA2}.
\paragraph{Other Data \& Preprocessing} We use the same preprocessing as well as development and test sets as in \S\ref{sec:a2}, except we 
learn $5$K \textsc{bpe}s as in \citet{wikimatrix}. 
\divfactors, \divagnostic, and \divtagged models are compared in controlled setups that use the same training data.
We also evaluate out-of-domain on the \texttt{khresmoi-summary} test set for the \textsc{wmt}$2014$ medical translation task~\cite{bojar-etal-2014-findings}.
\paragraph{Evaluation} We evaluate translation quality with \bleu~\cite{papineni-etal-2002-bleu} and \meteor~\cite{banerjee-lavie-2005-meteor}.\footnote{\url{https://github.com/mjpost/sacrebleu}}\textsuperscript{,}\footnote{\url{https://www.cs.cmu.edu/~alavie/METEOR/}} We compute Inference Expected Calibration Error (Inf\textsc{ece}) as ~\citet{Wang:2020:ACL}, which measures the difference in expectation between confidence and accuracy.\footnote{\url{https://github.com/shuo-git/InfECE}} We measure token-level translation accuracy based on Translation Error Rate (\textsc{ter}) alignments between hypotheses and references.\footnote{\url{http://www.cs.umd.edu/~snover/tercom/}} Unless mentioned otherwise, we decode with a beam size of $5$.

\subsection{Results}\label{sec:b3}
We discuss the impact of real divergences along the dimensions surfaced by the synthetic data analysis.

\paragraph{Translation Quality}
\begin{table*}[!ht]
    \centering
    \scalebox{0.8}{
    \begin{tabular}{ll@{\hskip 0.3in}l@{\hskip 0.3in}c@{\hskip 0.2in}c@{\hskip 0.5in}c@{\hskip 0.2in}r}
    
    & & &  \multicolumn{2}{c}{\textsc{fr}$\rightarrow$\textsc{en}}  & \multicolumn{2}{c}{\textsc{en}$\rightarrow$\textsc{fr}}  \\
    
    \addlinespace[0.2em]
    \cmidrule(lr){4-7}

    \multicolumn{2}{l}{\textsc{method}} & Training size &  \textsc{bleu} \ph  & \textsc{meteor} \ph  &   \textsc{bleu} \ph  & \textsc{meteor} \ph  \\
    
    \addlinespace[0.2em]
    \toprule

    \multicolumn{2}{l}{\textsc{laser}}   & \graydt{1.25} $1.25$M & $31.80$ {\small $\pm 0.36$} \ph & $34.00$ {\small $\pm 0.17$} \ph   & $32.16$ {\small $\pm 0.29$} \ph & $56.49$ {\small $\pm 0.24$}  \ph \\

    \addlinespace[0.2em]
    \arrayrulecolor{gray}\cmidrule(lr){1-7}
    \addlinespace[0.2em]

    \multicolumn{2}{l}{\textsc{equivalents}}  & \combdt{0.75}{0} $0.75$M   & \underline{$32.88$ {\small $\pm 0.07$}} \ph &  $34.75$ {\small $\pm 0.10$} \ph & $33.53$ {\small $\pm 0.35$} \ph & $57.38$ {\small $\pm 0.28$} \ph \\

    \addlinespace[0.2em]
    \addlinespace[0.2em]
    
    \multirow{3}{*}{\textsc{+div}\Bigg\{}  & \cellcolor{gray!5} \textsc{agnostic}  & \multirow{3}{*}{\combdt{0.75}{0.18} $0.93$M} & $32.47$ {\small $\pm 0.40$} \ph &  $34.56$ {\small $\pm 0.20$} \ph  & $33.19$ {\small $\pm 0.30$} \ph & $57.10$ {\small $\pm 0.30$} \ph \\
    & \cellcolor{gray!5} \textsc{tagged} & & $31.76$ {\small $\pm 1.61$} \ph & $34.17$ {\small $\pm 0.91$} \ph  & $\mathbf{33.43}$ {\small $\pm 0.39$} \ph & $\mathbf{57.55}$ {\small $\pm 0.27$} \ua\\
    & \cellcolor{gray!5} \textsc{factorized}  & & $32.73$ {\small $\pm 0.38$} \ph & $\mathbf{34.84}$ {\small $\pm 0.21$} \ua  & \underline{$\mathbf{33.92}$ {\small $\pm 0.38$}} \ua  & \underline{$\mathbf{57.63}$ {\small $\pm 0.28$}} \ua \\
    
    \addlinespace[0.2em]
    \addlinespace[0.2em]
    
    \multirow{3}{*}{\textsc{+div}\Bigg\{}  & \cellcolor{gray!5} \textsc{agnostic}  & \multirow{3}{*}{\combdt{0.75}{0.37} $1.12$M} & $32.53$ {\small $\pm 0.46$} \ph & $34.40$ {\small $\pm 0.21$} \ph & $31.47$ {\small $\pm 0.61$} \ph & $56.25$ {\small $\pm 0.46$} \ph \\
    & \cellcolor{gray!5} \textsc{tagged} & & $32.38$ {\small $\pm 0.40$} \ph  & $34.52$ {\small $\pm 0.13$} \ph  & $\mathbf{33.35}$ {\small $\pm 0.17$} \ua & $\mathbf{57.33}$ {\small $\pm 0.14$} \ua \\
    & \cellcolor{gray!5} \textsc{factorized}  & & $32.79$ {\small $\pm 0.24$} \ph & \underline{$\mathbf{34.89}$ {\small $\pm 0.12$}} \ua  & $\mathbf{33.22}$ {\small $\pm 0.35$} \ua & $\mathbf{57.31}$ {\small $\pm 0.30$} \ua\\

    \addlinespace[0.2em]
    \addlinespace[0.2em]
    
     \multirow{3}{*}{\textsc{+div}\Bigg\{}  & \cellcolor{gray!5} \textsc{agnostic} & \multirow{3}{*}{\combdt{0.75}{0.93} $1.68$M} & $31.40$ {\small $\pm 0.21$} \ph & $33.79$ {\small $\pm 0.11$} \ph  & $29.53$ {\small $\pm 0.39$} \ph & $54.29$ {\small $\pm 0.44$} \ph\\
     & \cellcolor{gray!5} \textsc{tagged}& & $31.97$ {\small $\pm 0.26$} \ua & $34.30$ {\small $\pm 0.10$} \ua & $31.37$ {\small $\pm 0.12$} \ua & $55.87$ {\small $\pm 0.18$} \ua\\
     & \cellcolor{gray!5} \textsc{factorized} & & $32.57$ {\small $\pm 0.19$} \ua  & $\mathbf{34.70}$ {\small $\pm 0.11$} \ua  & $31.60$ {\small $\pm 0.42$} \ua & $56.10$ {\small $\pm 0.22$} \ua\\

    \addlinespace[0.4em]
    \arrayrulecolor{black}\toprule
    \end{tabular}}
    \caption{
     Results for \textsc{en}$\leftrightarrow$\textsc{fr} translation on the \ted test set (averages and stdev of $3$ runs). We underline the top scores among all models and boldface the scores lying within one stdev from \equivalents. \ua \ \ denotes (one stdev) improvements of \divtagged and \divfactors over \divagnostic. Factorizing divergences helps \nmt recover from the degradation caused by divergences, while it achieves comparable scores to \equivalents.
    }
    \label{tab:main_results}
\end{table*}
\begin{table}[ht]
    \begin{center}
            \scalebox{0.73}{
    \begin{tabular}{ll@{\hskip 0.0in}cc}

    \textsc{method} & Train. size (M) & \fren & \enfr \\
    \addlinespace[0.4em]
    \toprule
    \addlinespace[0.5em]

    \laser & $1.25$ \graydt{1.25}                                   & $38.27$  {\small $\pm 0.49$}  &  $39.27$ {\small $\pm 0.45$} \\
    \equivalents & $0.75$ \combdt{0.75}{0.0}                        & $39.47$  {\small $\pm 0.24$}  &  $39.63$ {\small $\pm 0.52$} \\
    \addlinespace[0.8em]
    
    \multirow{3}{*}{\divagnostic}       & $0.93$ \combdt{0.75}{0.18} & $39.45$  {\small $\pm 0.50$} &  $39.78$ {\small $\pm 0.37$} \\ 
    & $1.12$ \combdt{0.75}{0.37}                                    & $\mathbf{40.00}$  {\small $\pm 0.14$} &  $39.20$ {\small $\pm 0.50$} \\
    & $1.68$ \combdt{0.75}{0.93}                                    & $\mathbf{39.90}$  {\small $\pm 0.14$} &  $38.00$ {\small $\pm 0.50$} \\
    
    \addlinespace[0.8em]
    
    \multirow{3}{*}{\divfactors} & $0.93$ \combdt{0.75}{0.18}   & \underline{$\mathbf{40.27}$   {\small $\pm 0.49$}} &  $40.13$ {\small $\pm 0.46$} \\
    & $1.12$ \combdt{0.75}{0.37}                                        & $\mathbf{40.03}$  {\small $\pm 0.42$} &  \underline{$\mathbf{40.30}$ {\small $\pm 0.29$}} \\
    & $1.68$ \combdt{0.75}{0.93}                                        & $\mathbf{39.97}$  {\small $\pm 0.26$} &  $39.30$ {\small $\pm 0.16$} \\
    
    \addlinespace[0.2em]
    \toprule
    \end{tabular}}
    \caption{\bleu scores on the medical domain. 
     We underline top scores and boldface (one stdev) improvements over \equivalents. Divergences improve translation quality when modeled by \divfactors.}\label{tab:medical_bleu}
    \end{center}
\end{table}
\begin{table}[!ht]
    \centering
    \scalebox{0.68}{
    \begin{tabular}{ccccccccccc}

    & \multicolumn{4}{c}{\textsc{training divergences}}\\
    \cmidrule(lr){2-5}
    \addlinespace[0.2em]
    \textsc{beam} & $0\%$ &  $20\%$ & $33\%$ & $55\%$ &\\
    \addlinespace[0.2em]
    \toprule
    \addlinespace[0.2em]
 
        \rowlabel{5.5}{1}{} & 
        \verticalchart{5.5}{1.93} & 
        \verticalpairchart{5.5}{1.55}{1.09} & 
        \verticalpairchart{5.5}{1.21}{1.16} & 
        \verticalpairchart{5.5}{2.92}{1.81}
        & \\
     
        \rowlabel{5.5}{5}{} & 
        \verticalchart{5.5}{1.53} & 
        \verticalpairchart{5.5}{1.19}{0.71} & 
        \verticalpairchart{5.5}{0.84}{0.84} & 
        \verticalpairchart{5.5}{2.78}{1.49}
        & \\    
        
        
        \rowlabel{5.5}{10}{} & 
        \verticalchart{5.5}{1.48} & 
        \verticalpairchart{5.5}{1.06}{0.73} & 
        \verticalpairchart{5.5}{0.84}{0.76} & 
        \verticalpairchart{5.5}{2.80}{1.28}
        & \\           

      \addlinespace[0.3em]
  
\end{tabular}}
\caption{Percentage of degenerated outputs for \fren models exposed to difference percentage of divergent training data (0\% corresponds to \textsc{equivalents}; dark gray columns correspond to \textsc{div-agnostic}).
\divfactors (grid-columns) help recover from degenerations, yielding fewer repetitions across beams.
}
\label{tab:degeneration_factorized}
\end{table}

Table~\ref{tab:main_results} presents \bleu and \meteor scores across model configurations and data settings on the \ted test sets. First, the model trained on \equivalents represents a very competitive baseline as it performs better or statistically comparable to all models.
This result is in line with prior evidence of \citet{vyas-etal-2018-identifying} who show that filtering out the most divergent pairs in noisy corpora (e.g., OpenSubtitles and CommonCrawl) does not hurt translation quality. 
Interestingly, the \equivalents model outperforms \laser
across metrics and translation directions, despite the fact that it is exposed to only about half of the training data. Gradually adding divergent data (\divagnostic) hurts translation quality across the board compared to the \equivalents model. The drops are significantly larger when divergences overwhelm the equivalent translations, which is consistent with our findings on synthetic data.

Second, \divfactors is the most effective mitigation strategy. With segment-level constraints (\divtagged), models can recover from the degradation caused by divergences (\divagnostic), but not consistently. 
By contrast, token-level factors (\divfactors) help \nmt recover from the impact of divergences across data setups and reach translation quality comparable to that of the \equivalents model, successfully mitigating the impact of the noisy training signals from divergent samples.

Third, when translating the out-of-domain test set, \divfactors improves over the \equivalents model, as presented in Table~\ref{tab:medical_bleu}. \divagnostic models perform comparably to \equivalents, while factorizing divergences improves on the latter by $\approx +1$ \bleu, for both directions.\footnote{We include \meteor results in Appendix~\ref{sec:appendix_meteor_medical}.} Mitigating the impact of divergences is thus important for \nmt to benefit from the increased coverage of out-of-domain data provided by the divergent samples. 

%
%
\paragraph{Degenerated Hypotheses} We check for degenerated outputs across models, data setups (we account for different percentages of divergences in the training data), and different beam sizes (Table~\ref{tab:degeneration_factorized}). As with synthetic divergences, we observe that when real divergences overwhelm the training data ($55\%$), degenerated loops are almost twice as frequent for all beam sizes. This phenomenon is consistently mitigated by \divfactors models across the board.\footnote{We observe similar trends for \enfr in Appendix~\ref{sec:en-fr-degenerations-appendix}} Furthermore, in some settings ($20\%$, $33\%$), \divfactors models decrease the amount of degenerated text by half compared to the \equivalents models.\footnote{\textsc{laser} models degenerate more frequently than \textsc{equivalents} and \textsc{div-factorized}.}  
%
\begin{figure*}[!ht]
\begin{subfigure}{.26\textwidth}
  \centering
  \includegraphics[width=\linewidth]{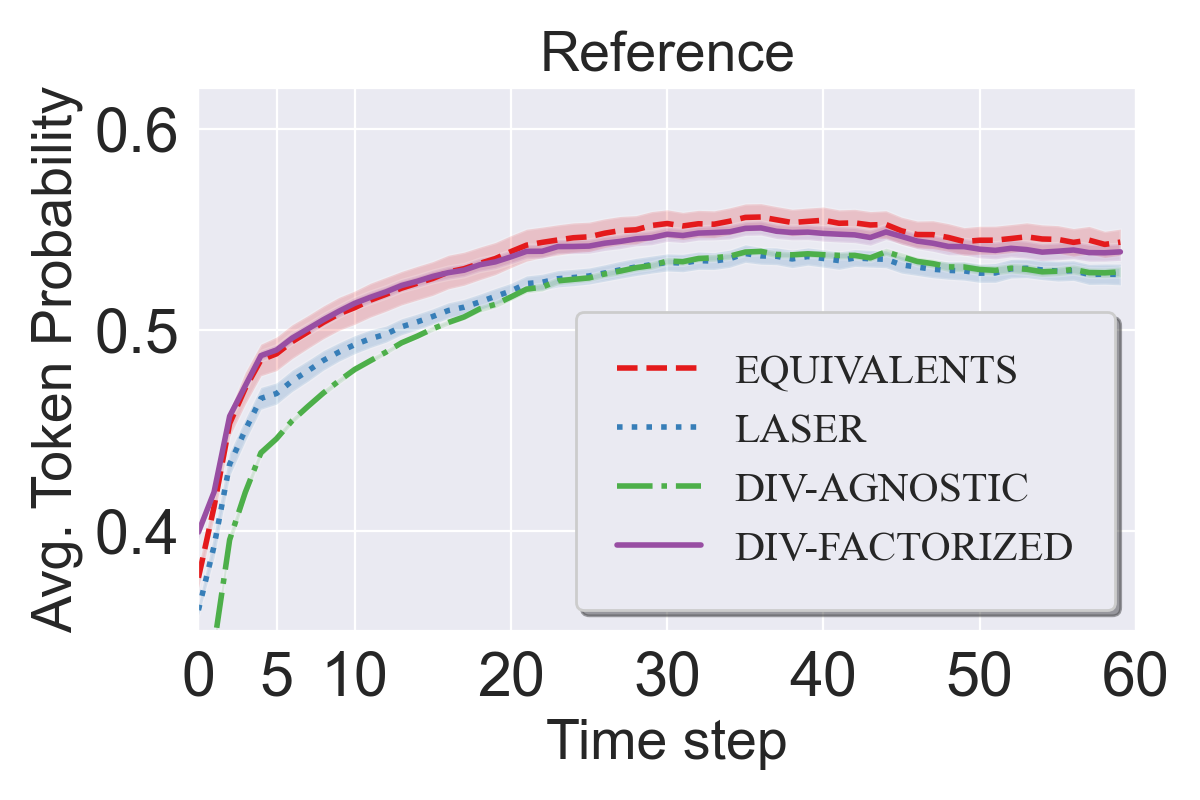}  \caption{\fren}
  \label{fig:avg_prob_fr2en_reference}
\end{subfigure}
\begin{subfigure}{.26\textwidth}
  \centering
  \includegraphics[width=\linewidth]{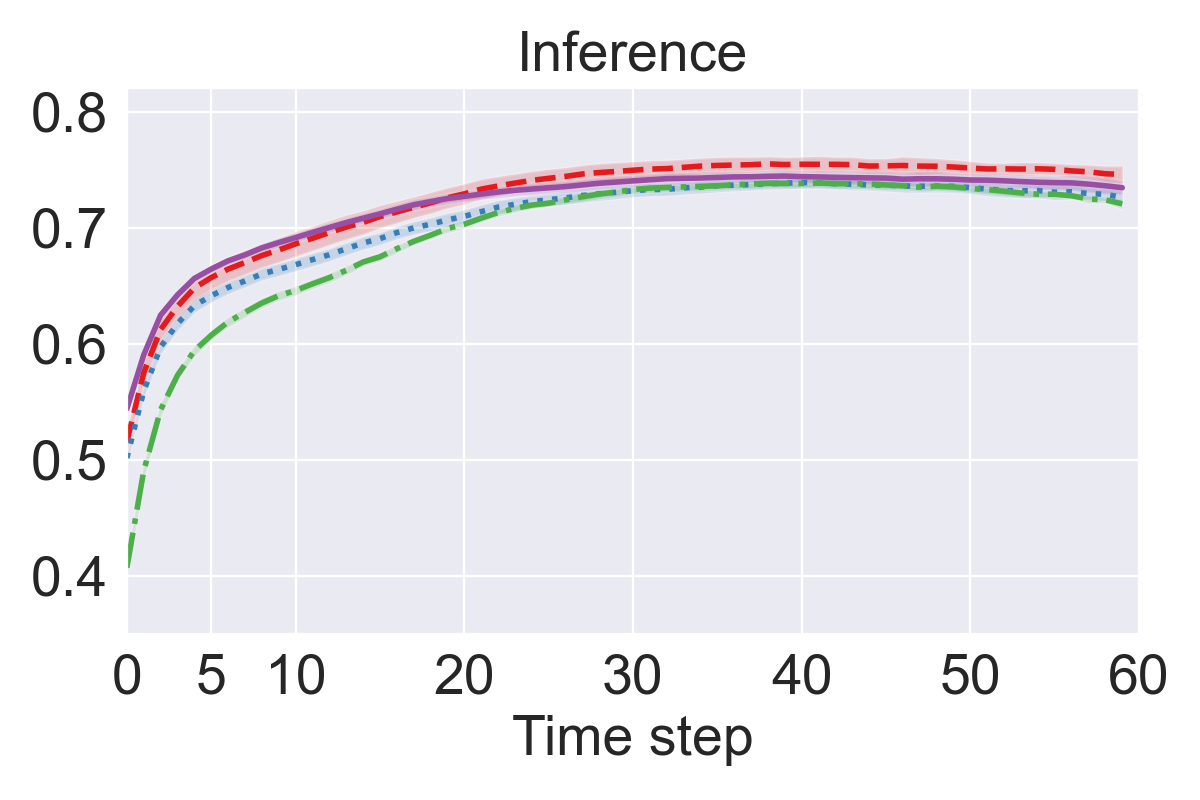}
  \caption{\fren}
  \label{fig:avg_prob_fr2en_inference}
\end{subfigure}%
\begin{subfigure}{.26\textwidth}
  \centering
  \includegraphics[width=\linewidth]{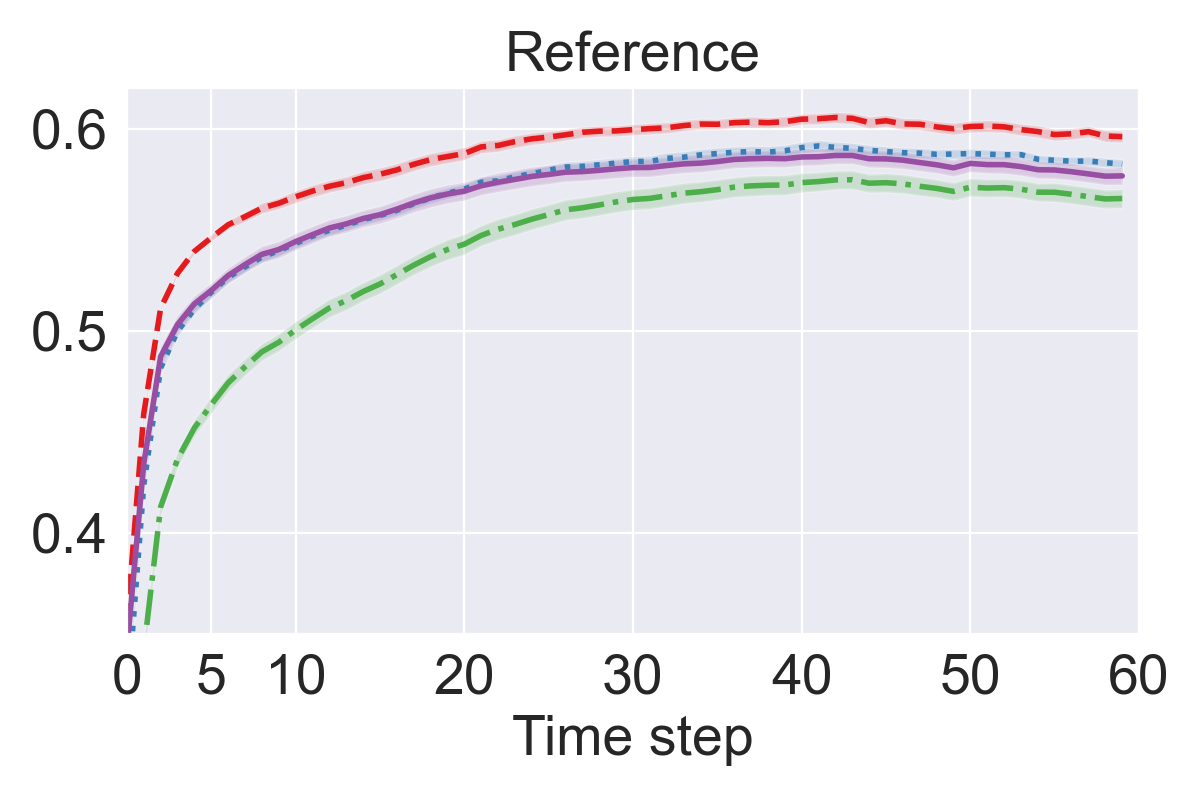}  \caption{\enfr}
  \label{fig:avg_prob_en2fr_reference}
\end{subfigure}%
\begin{subfigure}{.26\textwidth}
  \centering
  \includegraphics[width=\linewidth]{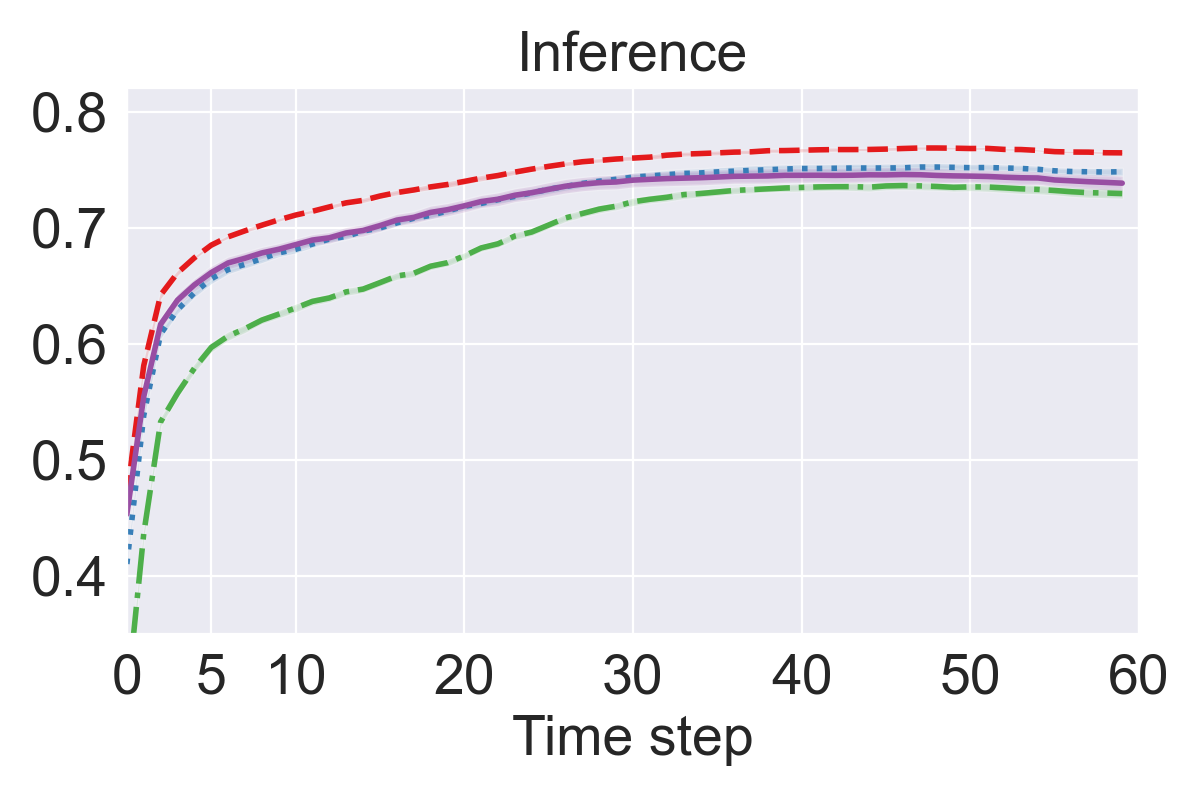}
  \caption{\enfr}
  \label{fig:avg_prob_en2fr_inference}
\end{subfigure}%
\caption{Average token probability across time steps on \ted test set.
\divagnostic yield the least confident predictions (for reference and inference prefixes); \divfactors help recover from this drop ($55\%$ divergences).} 
\label{fig:inference_cross_models_comparison_uncertainty}
\end{figure*}
\begin{table*}[!ht]
    \begin{center}
    \scalebox{0.75}{
    \begin{tabular}{l@{\hskip 0.2in}c@{\hskip 0.2in}l@{\hskip 0.2in}l@{\hskip 0.2in}l@{\hskip 0.3in}l@{\hskip 0.2in}l@{\hskip 0.2in}l}
    
     & &  \multicolumn{3}{c}{\textsc{fr}$\rightarrow$\textsc{en}} \ph \ph & \multicolumn{3}{c}{\textsc{en}$\rightarrow$\textsc{fr}} \ph \ph \\
    
    \addlinespace[0.2em]
    \cmidrule(lr){3-8}
    \addlinespace[0.2em]

    \textsc{method} & Train. size &  \textsc{Conf.} ($\uparrow$) & \textsc{Acc.} ($\uparrow$) & Inf\textsc{ece} ($\downarrow$) &  \textsc{Conf.\%} ($\uparrow$) & \textsc{Acc.} ($\uparrow$) & Inf\textsc{ece} ($\downarrow$) \\
    
    \addlinespace[0.2em]
    \toprule

    \textsc{laser}   & $1.25$M & $69.09$ {\small $\pm 0.67$} & $62.55$ {\small $\pm 0.29$} & $12.34$ {\small $\pm 0.38$} & $71.88$  {\small $\pm 0.30$} & $60.20$  {\small $\pm 0.18$} & $15.10$  {\small $\pm 0.12$}   \\

    \addlinespace[0.6em]

    \textsc{equivalents} & $0.75$M   & $70.96$ {\small $\pm 0.94$} & $63.49$ {\small $\pm 0.10$} & $12.37$ {\small $\pm 0.24$} & $74.35$  {\small $\pm 0.23$} & $61.81$  {\small $\pm 0.30$} & $15.09$  {\small $\pm 0.18$}\\

    \addlinespace[0.6em]

    \cellcolor{gray!5} \textsc{div-agnostic}  & \multirow{2}{*}{$0.93$M} &
    $71.19$ {\small $\pm 0.33$} & $63.54$ {\small $\pm 0.54$} & $\mathbf{12.00}$ {\small $\pm 0.06$} & $73.67$  {\small $\pm 0.11$} & $61.44$  {\small $\pm 0.22$} &  $15.19$  {\small $\pm 0.17$}\\
     \cellcolor{gray!5} \textsc{div-factorized}  & & \underline{$\mathbf{72.16}$ {\small $\pm 0.10$}}* & $\mathbf{64.29}$ {\small $\pm 0.44$}* & $\mathbf{11.81}$ {\small $\pm 0.04$}* & \underline{$74.50$  {\small $\pm 0.02$}}* & \underline{$\mathbf{62.26}$  {\small $\pm 0.27$}}* & \underline{$\mathbf{14.70}$  {\small $\pm 0.25$}}*\\
    
    \addlinespace[0.6em]

     \cellcolor{gray!5} \textsc{div-agnostic}  & \multirow{2}{*}{$1.12$M} & 
    $71.65$ {\small $\pm 0.18$} & $61.34$ {\small $\pm 0.33$} & $\mathbf{11.98}$ {\small $\pm 0.22$} & $71.72$  {\small $\pm 0.38$} & $59.29$  {\small $\pm 0.48$} & $15.62$  {\small $\pm 0.19$}\\
     \cellcolor{gray!5} \textsc{div-factorized}  & & $71.83$ {\small $\pm 0.03$} & \underline{$\mathbf{64.38}$ {\small $\pm 0.08$}}* & $\mathbf{11.86}$ {\small $\pm 0.01$} & $74.09$  {\small $\pm 0.14$}* &  $61.65$  {\small $\pm 0.19$}* & $\mathbf{14.84}$  {\small $\pm 0.18$}*\\

    \addlinespace[0.6em]

     \cellcolor{gray!5} \textsc{div-agnostic} & \multirow{2}{*}{$1.68$M} & 
     $68.01$ {\small $\pm 0.37$} & $61.34${ \small $\pm 0.23$} & $12.63$ {\small $\pm 0.23$} & $68.38$  {\small $\pm 0.25$} & $56.89$  {\small $\pm 0.34$} & $16.24$  {\small $\pm 0.27$}\\
     \cellcolor{gray!5} \textsc{div-factorized} & & $71.01$ {\small $\pm 0.39$}* & $63.65$ {\small $\pm 0.07$}* & \underline{$\mathbf{11.75}$ {\small $\pm 0.35$}}* & $71.81$  {\small $\pm 0.49$}* & $59.78$  {\small $\pm 0.39$}* & $14.95$  {\small $\pm 0.02$}*\\
     
    \addlinespace[0.4em]
    \arrayrulecolor{black}\toprule
    \end{tabular}}
    \caption{Average token confidence, accuracy, and inference calibration results for \textsc{en}$\leftrightarrow$\textsc{fr} translation on the \ted test set (average and stdev of $3$ runs). We underline top scores and boldface (one stdev) improvements over  \equivalents. * denotes (one stdev) improvements of \divfactors over \divagnostic. \textsc{div-factorized} yield more confident and accurate predictions compared to \divagnostic, yielding the smallest calibration errors.
    }\label{tab:b3_calibration_results_table}
    \end{center}
\end{table*}

\paragraph{Uncertainty} Figures~\ref{fig:avg_prob_fr2en_reference} and \ref{fig:avg_prob_en2fr_reference}
show that the gold-standard references are assigned lower probabilities by the \divagnostic models than all other models, especially in early time steps ($t<30$). We observe similar drops in confidence based on the probabilities of predicted tokens at inference time (\ref{fig:avg_prob_fr2en_inference} and \ref{fig:avg_prob_en2fr_inference}). This confirms that exposing models to fine-grained semantic divergences hurts their confidence, whether the divergences are synthetic or not. Furthermore, factorizing divergences helps mitigate the impact of naturally occurring divergences on uncertainty in addition to translation quality.

We conduct a calibration analysis to measure the differences between the confidence (i.e., \textit{probability}
) and the correctness  (i.e., \textit{accuracy}) of the generated tokens in expectation. 
Given that deep neural networks are often mis-calibrated in the direction of over-estimation (confidence$>$accuracy)~\cite{pmlr-v70-guo17a}, we check whether the increased confidence of \divfactors hurts calibration (Table~\ref{tab:b3_calibration_results_table}).  
\divfactors models are on average more confident \textit{and} more accurate than their \divagnostic counterparts. Interestingly, \divagnostic has smaller calibration errors than \equivalents and \laser models across the board.
%
%
%
\section{Related Work}\label{sec:factors_related_work}
We discuss work related to cross-lingual semantic divergences and noise effects in Section~\ref{sec:background_and_motivation} and now turn to the literature that connects with the methods used in this paper.
\paragraph{Factored Models} Factored models are introduced to inject word-level linguistic annotations (e.g., Part-of-Speech tags, lemmas) in translation. Source-side factors have been used in statistical \textsc{mt} \cite{fsmti} and in \nmt~\cite{sennrich-etal-2016-improving,hoang-etal-2016-improving}. Target-side factors are used by \citet{garcia2018} as an extension to the traditional \nmt framework that outputs multiple sequences. Although their main motivation is to enable models to handle larger vocabularies, \citet{fnmt-appli} propose a list of novel applications of target-side factors beyond their initial purpose, such as word-case prediction and subword segmentation. Our approach draws inspiration from all the aforementioned works, yet it is unique in its use of \textit{both} source and target factors to incorporate \textit{semantics} in \nmt.  
\paragraph{Calibration} 
\citet{kumar} 
find that \nmt models are miscalibrated, even when conditioned on gold-standard prefixes. They attribute this behavior to the poor calibration of the \textsc{eos} token and the uncertainty of attention and design a recalibration model to improve calibration. \citet{ott2018analyzing} argue that miscalibration can be attributed to the ``extrinsic'' uncertainty of the noisy, untranslated references found in the training data. \citet{muller} investigate the effect of label smoothing on calibration. 
On a similar spirit, \citet{Wang:2020:ACL} propose graduated label smoothing to improve calibration at inference time. They also link miscalibration to linguistic properties of the data (e.g., frequency, position, syntactic roles). Our work, in contrast, focuses on the semantic properties of the training data that affect calibration.

\label{sec:related_work}
%
%
\section{Conclusion}\label{sec:conclusion}

This work investigates the impact of semantic mismatches beyond noise in parallel text on \nmt quality and confidence. Our experiments on \textsc{en}$\leftrightarrow$\textsc{fr} tasks show that fine-grained semantic divergences hurt translation quality when they overwhelm the training data. Models exposed to fine-grained divergences at training time are less confident in their predictions, which hurts beam search and produces degenerated text (repetitive loops) more frequently. 

Furthermore, we also show that, unlike noisy samples, fine-grained divergences can still provide a useful training signal for \nmt when they are modeled via factors. Evaluated on \textsc{en}$\leftrightarrow$\textsc{fr} translation tasks, our divergent-aware \nmt framework mitigates the negative impact of divergent references on translation quality, improves the confidence and calibration of predictions, and produces degenerated text less frequently.

More broadly, this work illustrates how understanding the properties of training data can help build better \nmt models. In future work, we will extend our analysis to other properties of parallel text and to other language pairs, focusing on low-resource conditions where divergences are expected to be even more prevalent. 

\section*{Acknowledgements}

We thank Sweta Agrawal, Doug Oard, 
Suraj Rajappan Nair,
the anonymous reviewers and the \textsc{clip} lab at \textsc{umd} for helpful comments. This material is based upon work supported by the National Science Foundation under Award No.\ $1750695$. Any opinions, findings, and conclusions or recommendations expressed in this material are those of the authors and do not necessarily reflect the views of the National Science Foundation.
%
%
\bibliographystyle{acl_natbib}
\bibliography{anthology,acl2021}
%
%
\clearpage
\appendix
\section{WikiMatrix Fine-grained Divergences}\label{sec:appendixA1}
Starting from the pool of examples identified as divergent under the divergentm\textsc{bert} classifier, we want to focus on the subset of samples that contain fine meaning differences. Therefore, we use \texttt{bicleaner} to filter out training data that are likely to contain coarse meaning differences. \citet{espla-gomis-etal-2020-bicleaner} report better \nmt results on English$\leftrightarrow$Portuguese translation after cleaning WikiMatrix data with thresholds of $0.5$ and $0.7$.

\begin{figure}[!ht]
\begin{center}
\begin{subfigure}{0.5\textwidth}
  \centering
  \includegraphics[scale=0.4]{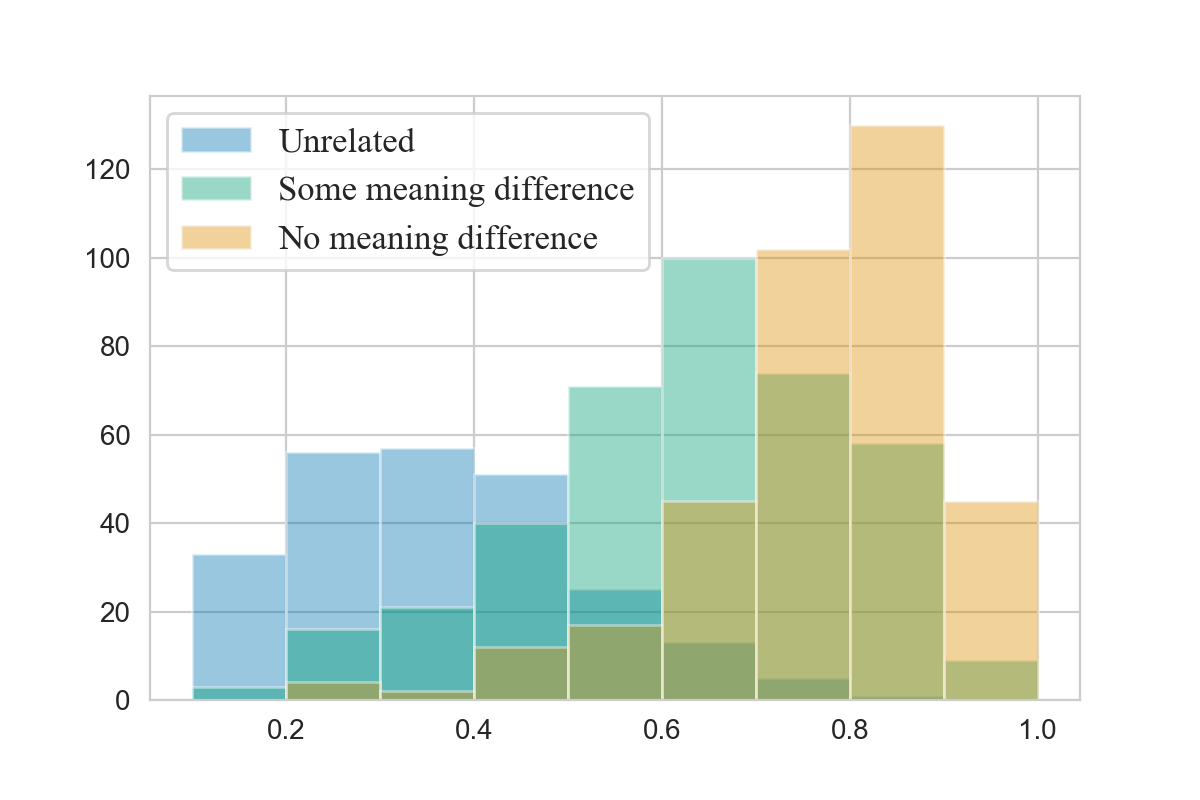}
   \caption{\textsc{refresd}}
\end{subfigure}
\begin{subfigure}{0.5\textwidth}
  \centering
  \includegraphics[scale=0.4]{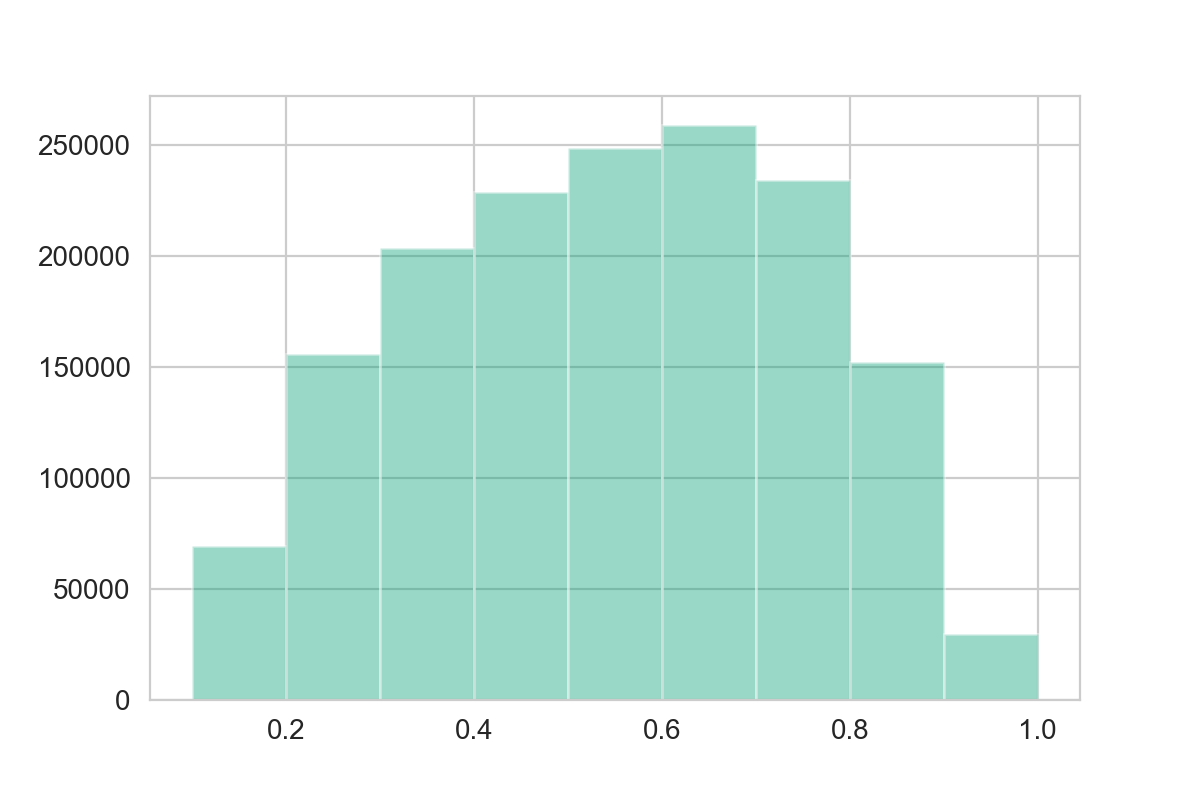}
   \caption{\textsc{WikiMatrix divergences}}
\end{subfigure}
\caption{Distribution of bicleaner score on \textsc{refresd}, and English-French WikiMatrix divergences.}\label{fig:bicleaner}
\end{center}
\end{figure}

We conduct a preliminary experiment to understand how the \texttt{bicleaner} scores of English-French WikiMatrix sentences are distributed. Figure~\ref{fig:bicleaner}(a) 
shows the distribution of scores among the three classes of the \textsc{refresd} dataset, a dataset that distinguishes fine meaning differences (``some meaning difference''), coarse divergences (``unrelated''), and equivalent translation pairs (``no meaning difference'').\footnote{\url{https://github.com/Elbria/xling-SemDiv/tree/master/REFreSD}}
We observe that thresholding the \texttt{bicleaner} score at $>0.5$ filters out most of the unrelated pairs. We conduct three experiments with thresholds at $0.8$, $0.7$, and $0.5$ to gradually add more fine-grained divergences. Figure~\ref{fig:bicleaner}(b) presents the number of English-French WikiMatrix divergences, binned by \texttt{bicleaner} scores.
%
\section{\texttt{Sockeye2} configuration details}\label{sec:appendixA2}
Tables~\ref{tab:sockeye_main_config} and \ref{tab:sockeye_factors_config} present details of \nmt training with~\texttt{Sockeye2}.
\begin{table}[!ht]
    \centering
    \scalebox{0.9}{
    \begin{tabular}{|l|}
    \hline
    \texttt{--weight-tying-type="trg\_softmax"} \\
    \texttt{--num-words $5000$:$5000$} \\
    \texttt{--label-smoothing $0.1$} \\
    \texttt{--encoder transformer} \\
    \texttt{--decoder transformer} \\
    \texttt{--num-layers $6$} \\
    \texttt{--transformer-attention-heads $84$} \\
    \texttt{--transformer-model-size $512$} \\
    \texttt{--num-embed $512$} \\
    \texttt{--transformer-feed-forward-num-hidden $2048$} \\
    \texttt{--transformer-preprocess n} \\
    \texttt{--transformer-postprocess dr} \\
    \texttt{--gradient-clipping-type none} \\
    \texttt{--transformer-dropout-attention $0.1$} \\
    \texttt{--transformer-dropout-act $0.1$} \\
    \texttt{--transformer-dropout-prepost $0.1$} \\
    \texttt{--max-seq-len $80$:$80$}\\
    \texttt{--batch-type word} \\
    \texttt{--batch-size $2048$} \\
    \texttt{--min-num-epochs $3$} \\
    \texttt{--initial-learning-rate $0.0002$} \\
    \texttt{--learning-rate-reduce-factor $0.7$} \\
    \texttt{--learning-rate-reduce-num-not-improved $4$} \\
    \texttt{--checkpoint-interval $1000$} \\
    \texttt{--keep-last-params $30$} \\
    \texttt{--max-num-checkpoint-not-improved $20$} \\
    \texttt{--decode-and-evaluate $1000$} \\
    \hline
    \end{tabular}}\vspace{-0.5em}
    \caption{\nmt configurations on \texttt{Sockeye2} for \equivalents, \laser, \divagnostic, and \divtagged.}
    \label{tab:sockeye_main_config}
\end{table}
\begin{table}[!ht]
    \centering
    \scalebox{0.9}{
    \begin{tabular}{|l@{\hskip 0.9in}|}
    \hline
    \texttt{--weight-tying-type none} \\
    \texttt{--source-factors-num-embed $8$} \\
    \texttt{--source-factors-combine concat} \\
    \texttt{--target-factors-num-embed $8$} \\
    \texttt{--target-factors-combine concat} \\
    \texttt{--transformer-model-size $504$:$512$} \\
    \texttt{--num-embed $504$:$504$} \\
    \hline
    \end{tabular}}
    \caption{\nmt configurations on \texttt{Sockeye2} for \divfactors; for missing settings refer to Table~\ref{tab:sockeye_main_config}.}
    \label{tab:sockeye_factors_config}
\end{table}\vspace{-1em}
\section{Measuring Degenerated Hypotheses}\label{sec:AppendixA3}
We include the pseudo-algorithm that checks if a hypothesis falls under
odd repetitions not supported by the reference in Algorithm~\ref{alg:degenerated_test}. When measuring repeated $n$-grams we exclude punctuation and conjunctions. The \textsc{Repeated} function checks whether an $n$-gram is repeated (number of occurrences $>1$) in the hypothesis $h$, or reference $r$.
\begin{table*}[!t]
    \centering
    \scalebox{0.9}{
    \begin{tabular}{lccccr}
    
    \toprule
    \addlinespace[0.1em]
    WikiMatrix version & \#Sents. & \#Tokens & \#Types & Length & \%Corr\\
    
    \addlinespace[0.1em]
    \cmidrule{2-6}
    \addlinespace[0.1em]
    
    \equivalents  & $751{,}792$ & $22{,}723{,}543$  & $515{,}154$ & $30.2$ &  $0\%$\\
    \subtree      & $749{,}973$ &  $20{,}783{,}056$ & $483{,}336$ & $27.7$ &  $9.32\%$\\
    \replacement  & $750{,}527$ & $22{,}735{,}143$  & $475{,}567$ & $30.3$ &  $16.11\%$\\
    \lexical (\textsc{hypernyms}) & $724{,}326$  & $22{,}014{,}609$  & $497{,}658$ & $30.4$ &  $12.33\%$\\
    \lexical (\textsc{hyponyms}) & $617{,}913$ & $18{,}970{,}039$   & $442{,}299$ & $30.7$ &  $7.42\%$\\
    
    \addlinespace[0.1em]
    \toprule
    \end{tabular}}
    \caption{WikiMatrix statistics corresponding to extracted \equivalents and the fine-grained corruptions introduced in the synthetic setting (\textsc{en}-side). $\%$Corr denotes the average $\%$ of corrupted tokens in a sentence.}
    \label{tab:synthetic_statistics_english}
\end{table*}
\begin{table*}[!t]
    \centering
    \scalebox{0.9}{
    \begin{tabular}{lccccr}
    
    \toprule
    \addlinespace[0.1em]
    
    WikiMatrix Version & \#Sents. & \#Tokens & \#Types & Length & \%Corr\\
    
    \addlinespace[0.1em]
    \cmidrule{2-6}
    \addlinespace[0.1em]
    
    \equivalents  & $751{,}792$ & $25{,}554{,}549$  & $515{,}194$ & $34.0$ & $0\%$  \\
    \subtree      & $749{,}973$ & $23{,}822{,}958$  & $486{,}908$ & $31.8$ &  $7.21\%$\\
    \replacement  & $750{,}527$ & $25{,}554{,}549$  & $515{,}194$ & $34.0$ & $12.74\%$ \\
    \lexical (\textsc{hypernyms}) & $724{,}326$  & $24{,}737{,}604$   & $499{,}423$ & $34.2$ & $9.82\%$ \\
    \lexical (\textsc{hyponyms}) & $617{,}913$ & $21{,}387{,}650$  & $445{,}871$ & $34.6$ & $5.78\%$ \\
    
    \addlinespace[0.1em]
    \toprule
    \end{tabular}}
    \caption{WikiMatrix statistics corresponding to extracted \equivalents and the fine-grained corruptions introduced in the synthetic setting (\textsc{fr}-side).}
    \label{tab:synthetic_statistics_french}
\end{table*}
\begin{algorithm}[!ht] 
\caption{Degenerated hypothesis check}
\label{alg:loop}
\begin{algorithmic}[1]
\Require{$h$, $r$ (hypothesis, reference)} 
\Ensure{$Deg$ (True for degenerated hypothesis)}
\Statex
\Function{DegenerationCheck}{$h[\;]$,$r[\;]$}
    \For{$n$-gram $\in$ $h$}   
        \If{\textsc{Repeated}($n$-gram, $h$)}  
            \If{ not \textsc{Repeated}($n$-gram, $r$)}
                \State \Return True
            \EndIf
        \EndIf
    \EndFor
    \State \Return False
\EndFunction
\end{algorithmic}\label{alg:degenerated_test}
\end{algorithm}\vspace{-1em}
\section{Synthetic Divergences Statistics}\label{sec:synthetic_divergences_stats}
Tables~\ref{tab:synthetic_statistics_english} and \ref{tab:synthetic_statistics_french} contain corpus statistics for the $3$ versions of synthetic divergences we create, starting from \equivalents. \lexical are sampled at random from the pools of substitutions based on hypernyms and hyponyms. 
\begin{table}[!t]
    \centering
    \scalebox{0.7}{
    \begin{tabular}{ll@{\hskip 0.2in}cc}

    \addlinespace[0.2em]

    \textsc{method} & Training data (M)  & \fren & \enfr \\
    
    \addlinespace[0.2em]
    \toprule
    \addlinespace[0.2em]

    \laser & $1.25$ \graydt{1.25}                                    & $40.67$ {\small $\pm 0.28$}  & $65.17$ {\small $\pm 0.40$}\\
    \equivalents & $0.75$ \combdt{0.75}{0.0}                         & $41.23$ {\small $\pm 0.16$}   & $65.60$ {\small $\pm 0.45$}\\
    \addlinespace[0.5em]
    
    \multirow{3}{*}{\divagnostic} & $0.93$ \combdt{0.75}{0.18}  & $41.25$ {\small $\pm 0.17$} & $65.67$ {\small $\pm 0.22$}\\ 
    & $1.12$ \combdt{0.75}{0.37}                                     & $41.19$ {\small $\pm 0.31$}  & $65.23$ {\small $\pm 0.33$}\\
    & $1.68$ \combdt{0.75}{0.93}                                     & $41.07$ {\small $\pm 0.13$}  & $63.97$ {\small $\pm 0.56$}\\
    
    \addlinespace[0.5em]
    
    \multirow{3}{*}{\divfactors} & $0.93$ \combdt{0.75}{0.18}    &  $41.34$ {\small $\pm 0.23$} & \underline{$66.03$ {\small $\pm 0.38$}}\\
    & $1.12$ \combdt{0.75}{0.37}                                        &  \underline{$41.33$ {\small $\pm 0.31$}} &  $65.88$ {\small $\pm 0.35$}\\
    & $1.68$ \combdt{0.75}{0.93}                                        &  $41.25$ {\small $\pm 0.08$} &  $65.08$ {\small $\pm 0.11$}\\
    
    \addlinespace[0.2em]
    \toprule
    \end{tabular}}
    \caption{\meteor scores on medical translation task.}
    \label{tab:medical_meteor}\vspace{-1em}
\end{table}
\section{\meteor Results (addition)}\label{sec:appendix_meteor_medical}
For completeness, we present \meteor scores to complement the \bleu evaluation of \S\ref{sec:b3}, which consists the official evaluation metric of \textsc{wmt} biomedical translation tasks~\cite{jimeno-yepes-etal-2017-findings,neves-etal-2018-findings,bawden-etal-2019-findings,bawden-etal-2020-findings}. The average improvements of \divfactors over \equivalents and \divagnostic are smaller compared to the differences highlighted by \bleu. However, we note that \meteor results might be misleading when evaluating medical translations, as in this domain we might not want to account for synonyms when comparing references to hypotheses. 
\section{Degenerated Hypotheses (addition)}\label{sec:en-fr-degenerations-appendix}
\divfactors decreases the $\%$ of degenerated outputs caused by divergent data (Table~\ref{tab:degeneration_factorized_en2fr}).
\begin{table}[!ht]
    \centering
    \scalebox{0.57}{
    \begin{tabular}{ccccccccccc}

    & \multicolumn{4}{c}{\textsc{divergences}}\\
    \cmidrule(lr){2-5}
    \addlinespace[0.2em]
    \textsc{beam} & $0\%$ &  $20\%$ & $33\%$ & $55\%$ &\\
    \addlinespace[0.2em]
    \toprule
    \addlinespace[0.2em]
    
        \rowlabel{5.5}{1}{} & 
        \verticalchart{5.5}{1.41} & 
        \verticalpairchart{5.5}{1.95}{1.15} & 
        \verticalpairchart{5.5}{2.09}{1.85} & 
        \verticalpairchart{5.5}{2.24}{2.17}
        & \\  
 
        \rowlabel{5.5}{5}{} & 
        \verticalchart{5.5}{1.10} & 
        \verticalpairchart{5.5}{1.19}{0.71} & 
        \verticalpairchart{5.5}{1.87}{1.29} & 
        \verticalpairchart{5.5}{2.30}{1.59}
        & \\    

        \rowlabel{5.5}{10}{} & 
        \verticalchart{5.5}{0.89} & 
        \verticalpairchart{5.5}{1.21}{0.65} & 
        \verticalpairchart{5.5}{1.78}{1.10} & 
        \verticalpairchart{5.5}{2.30}{1.80}
        & \\    
      \addlinespace[0.3em]
\end{tabular}}
\caption{$\%$ of degenerated outputs across beams (\enfr). \divfactors (grid-columns) help recover from degenerations, yielding fewer repetitions.}
\label{tab:degeneration_factorized_en2fr}
\end{table}

\end{document}